\documentclass{article}

\usepackage[nonatbib, preprint]{neurips_2018}

\usepackage[utf8]{inputenc} % allow utf-8 input
\usepackage[T1]{fontenc}    % use 8-bit T1 fonts
\usepackage{hyperref}
\usepackage{url}            % simple URL typesetting
\usepackage{booktabs}       % professional-quality tables
\usepackage{amsfonts}       % blackboard math symbols
\usepackage{nicefrac}       % compact symbols for 1/2, etc.
\usepackage{microtype}      % microtypography
\usepackage{graphicx}
\usepackage{mathtools}
\usepackage{amsthm}

\title{From GAN to WGAN}

\author{%
  Lilian~Weng\\
  OpenAI \\
  \texttt{lilian@openai.com}
}

\begin{document}

\maketitle

\begin{abstract}
This paper explains the math behind a generative adversarial network (GAN)~\cite{gan2014} model and why it is hard to be trained. Wasserstein GAN is intended to improve GANs' training by adopting a smooth metric for measuring the distance between two probability distributions.
\end{abstract}

\section{Introduction}

Generative adversarial network (GAN)~\cite{gan2014} has shown great results in many generative tasks to replicate the real-world rich content such as images, human language, and music. It is inspired by game theory: two models, a generator and a critic, are competing with each other while making each other stronger at the same time. However, it is rather challenging to train a GAN model, as people are facing issues like training instability or failure to converge. 

Here I would like to explain the math behind the generative adversarial network framework,  why it is hard to be trained, and finally introduce a modified version of GAN intended to solve the training difficulties.

\section{Kullback–Leibler and Jensen–Shannon Divergence}
\label{sec:kl_and_js}

Before we start examining GANs closely, let us first review two metrics for quantifying the similarity between two probability distributions.

(1) \textbf{KL (Kullback–Leibler) Divergence} measures how one probability distribution $p$ diverges from a second expected probability distribution $q$.

\[
D_{KL}(p \| q) = \int_x p(x) \log \frac{p(x)}{q(x)} dx
\]

$D_{KL}$ achieves the minimum zero when $p(x) == q(x)$ everywhere.

It is noticeable according to the formula that KL divergence is asymmetric. In cases where $p(x)$ is close to zero, but $q(x)$ is significantly non-zero, the $q$'s effect is disregarded. It could cause buggy results when we just want to measure the similarity between two equally important distributions.

(2) \textbf{Jensen–Shannon Divergence} is another measure of similarity between two probability distributions, bounded by $[0, 1]$. JS divergence is symmetric and more smooth. Check this \href{https://www.quora.com/Why-isnt-the-Jensen-Shannon-divergence-used-more-often-than-the-Kullback-Leibler-since-JS-is-symmetric-thus-possibly-a-better-indicator-of-distance}{post} if you are interested in reading more about the comparison between KL divergence and JS divergence.

\[
D_{JS}(p \| q) = \frac{1}{2} D_{KL}(p \| \frac{p + q}{2}) + \frac{1}{2} D_{KL}(q \| \frac{p + q}{2})
\]

\begin{figure}[!htb]
	\centering
	\includegraphics[width=\linewidth]{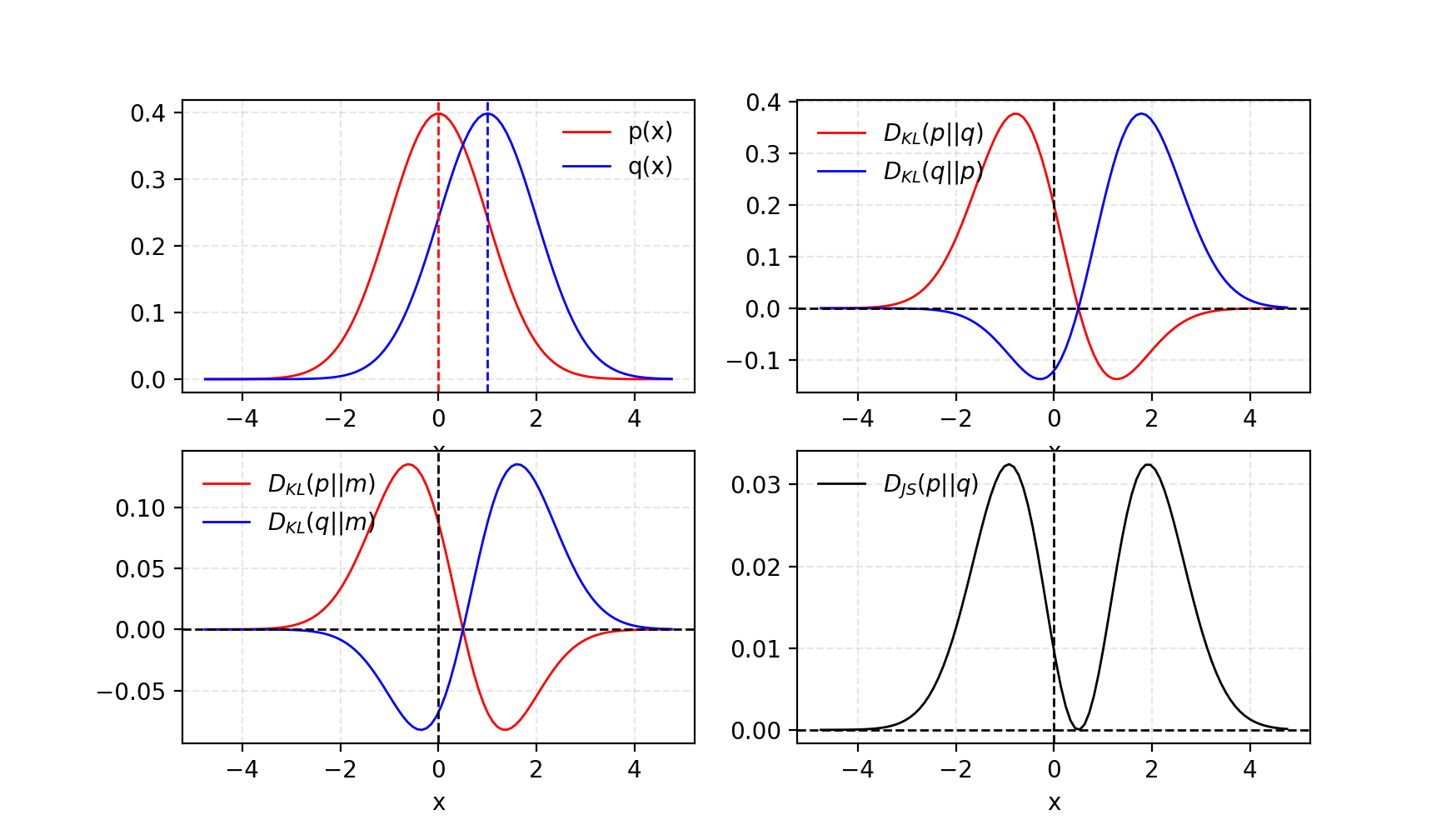}
	\caption{Given two Gaussian distribution, $p$ with mean=0 and std=1 and $q$ with mean=1 and std=1. The average of two distributions is labeled as $m=(p+q)/2$. KL divergence $D_{KL}$ is asymmetric but JS divergence $D_{JS}$ is symmetric.}
	\label{fig:fig1}
\end{figure}

Some~\cite{gan2015train} believe that one reason behind GANs' big success is switching the loss function from asymmetric KL divergence in traditional maximum-likelihood approach to symmetric JS divergence. We will discuss more on this point in the next section.

\section{Generative Adversarial Network}

GAN consists of two models:
\begin{itemize}
    \item A discriminator $D$ estimates the probability of a given sample coming from the real dataset. It works as a critic and is optimized to tell the fake samples from the real ones.
	\item A generator $G$ outputs synthetic samples given a noise variable input $z$ ($z$ brings in potential output diversity). It is trained to capture the real data distribution so that its generative samples can be as real as possible, or in other words, can trick the discriminator to offer a high probability.
\end{itemize}

\begin{figure}[!htb]
	\centering
	\includegraphics[width=0.85\linewidth]{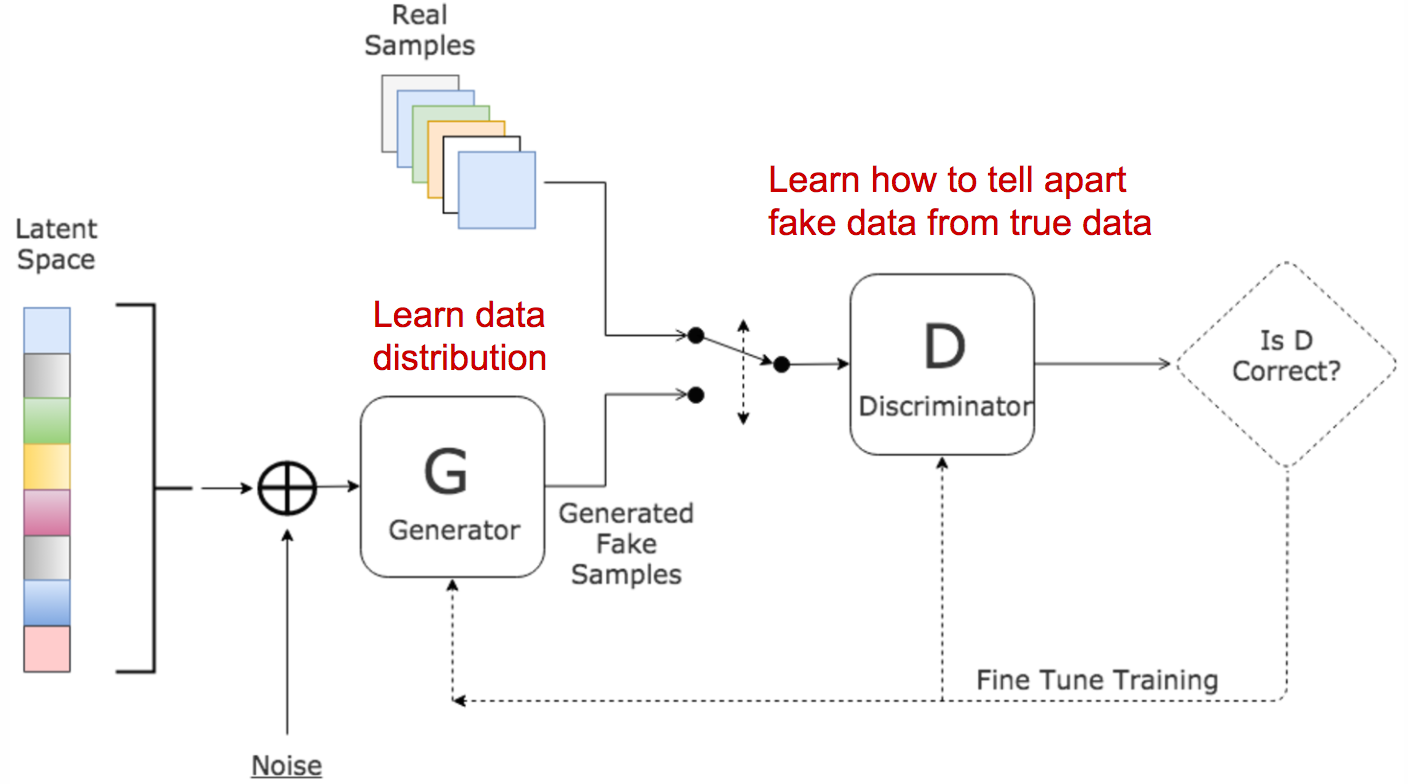}
	\caption{Architecture of a generative adversarial network. (Image source: \href{http://www.kdnuggets.com/2017/01/generative-adversarial-networks-hot-topic-machine-learning.html}{KDNuggets}).}
	\label{fig:fig2}
\end{figure}

These two models compete against each other during the training process: the generator $G$ is trying hard to trick the discriminator, while the critic model $D$ is trying hard not to be cheated. This interesting zero-sum game between two models motivates both to improve their functionalities.

Given, 

\begin{table}[h!]
	\centering
	\begin{tabular}{c|l|l} 
		\hline
		\textbf{Symbol} & \textbf{Meaning} & \textbf{Notes}\\
		\hline
		$p_{z}$ & Data distribution over noise input $z$ & Usually, just uniform. \\
		$p_{g}$ & The generator's distribution over data $x$ & \\
		$p_{r}$ &  Data distribution over real sample $x$ & \\
		\hline
	\end{tabular}
\end{table}

On one hand, we want to make sure the discriminator $D$'s decisions over real data are accurate by maximizing $\mathbb{E}_{x \sim p_{r}(x)} [\log D(x)]$. Meanwhile, given a fake sample $G(z), z \sim p_z(z)$, the discriminator is expected to output a probability, $D(G(z))$, close to zero by maximizing $\mathbb{E}_{z \sim p_{z}(z)} [\log (1 - D(G(z)))]$.

On the other hand, the generator is trained to increase the chances of $D$ producing a high probability for a fake example, thus to minimize $\mathbb{E}_{z \sim p_{z}(z)} [\log (1 - D(G(z)))]$.

When combining both aspects together, $D$ and $G$ are playing a \textit{minimax game} in which we should optimize the following loss function:

\[
\begin{aligned}
\min_G \max_D L(D, G) 
& = \mathbb{E}_{x \sim p_{r}(x)} [\log D(x)] + \mathbb{E}_{z \sim p_z(z)} [\log(1 - D(G(z)))] \\
& = \mathbb{E}_{x \sim p_{r}(x)} [\log D(x)] + \mathbb{E}_{x \sim p_g(x)} [\log(1 - D(x)]
\end{aligned}
\]

where $\mathbb{E}_{x \sim p_{r}(x)} [\log D(x)]$ has no impact on $G$ during gradient descent updates.

\subsection{What is the Optimal Value for D?}

Now we have a well-defined loss function. Let's first examine what is the best value for $D$.

\[
L(G, D) = \int_x \bigg( p_{r}(x) \log(D(x)) + p_g (x) \log(1 - D(x)) \bigg) dx
\]

Since we are interested in what is the best value of $D(x)$ to maximize $L(G, D)$, let us label 

\[
\tilde{x} = D(x), 
A=p_{r}(x), 
B=p_g(x)
\]

And then what is inside the integral (we can safely ignore the integral because $x$ is sampled over all the possible values) is:

\begin{align*}
f(\tilde{x}) 
& = A log\tilde{x} + B log(1-\tilde{x}) \\
\frac{d f(\tilde{x})}{d \tilde{x}}
& = A \frac{1}{ln10} \frac{1}{\tilde{x}} - B \frac{1}{ln10} \frac{1}{1 - \tilde{x}} \\
& = \frac{1}{ln10} (\frac{A}{\tilde{x}} - \frac{B}{1-\tilde{x}}) \\
& = \frac{1}{ln10} \frac{A - (A + B)\tilde{x}}{\tilde{x} (1 - \tilde{x})} \\
\end{align*}

Thus, set $\frac{d f(\tilde{x})}{d \tilde{x}} = 0$, we get the best value of the discriminator: $D^*(x) = \tilde{x}^* = \frac{A}{A + B} = \frac{p_{r}(x)}{p_{r}(x) + p_g(x)} \in [0, 1]$.
Once the generator is trained to its optimal, $p_g$ gets very close to $p_{r}$. When $p_g = p_{r}$, $D^*(x)$ becomes $1/2$.

\subsection{What is the Global Optimal? }

When both $G$ and $D$ are at their optimal values, we have $p_g = p_{r}$ and $D^*(x) = 1/2$ and the loss function becomes:

\begin{align*}
L(G, D^*) 
&= \int_x \bigg( p_{r}(x) \log(D^*(x)) + p_g (x) \log(1 - D^*(x)) \bigg) dx \\
&= \log \frac{1}{2} \int_x p_{r}(x) dx + \log \frac{1}{2} \int_x p_g(x) dx \\
&= -2\log2
\end{align*}

\subsection{What does the Loss Function Represent?}

According to the formula listed in Sec.~\ref{sec:kl_and_js}, JS divergence between $p_{r}$ and $p_g$ can be computed as:
	
\begin{align*}
D_{JS}(p_{r} \| p_g) 
=& \frac{1}{2} D_{KL}(p_{r} || \frac{p_{r} + p_g}{2}) + \frac{1}{2} D_{KL}(p_{g} || \frac{p_{r} + p_g}{2}) \\
=& \frac{1}{2} \bigg( \log2 + \int_x p_{r}(x) \log \frac{p_{r}(x)}{p_{r} + p_g(x)} dx \bigg) + \\& \frac{1}{2} \bigg( \log2 + \int_x p_g(x) \log \frac{p_g(x)}{p_{r} + p_g(x)} dx \bigg) \\
=& \frac{1}{2} \bigg( \log4 + L(G, D^*) \bigg)
\end{align*}

Thus, 

\[
L(G, D^*) = 2D_{JS}(p_{r} \| p_g) - 2\log2
\]

Essentially the loss function of GAN quantifies the similarity between the generative data distribution $p_g$ and the real sample distribution $p_{r}$ by JS divergence when the discriminator is optimal. The best $G^*$ that replicates the real data distribution leads to the minimum $L(G^*, D^*) = -2\log2$ which is aligned with equations above.

\textbf{Other Variations of GAN}: There are many variations of GANs in different contexts or designed for different tasks. For example, for semi-supervised learning, one idea is to update the discriminator to output real class labels, $1, \dots, K-1$, as well as one fake class label $K$. The generator model aims to trick the discriminator to output a classification label smaller than $K$.

\section{Problems in GANs}

Although GAN has shown great success in the realistic image generation, the training is not easy; The process is known to be slow and unstable.

\subsection{Hard to Achieve Nash Equilibrium}

\cite{salimans2016nips} discussed the problem with GAN's gradient-descent-based training procedure. Two models are trained simultaneously to find a Nash equilibrium to a two-player non-cooperative game. However, each model updates its cost independently with no respect to another player in the game. Updating the gradient of both models concurrently cannot guarantee a convergence.

Let's check out a simple example to better understand why it is difficult to find a Nash equilibrium in an non-cooperative game. Suppose one player takes control of $x$ to minimize $f_1(x) = xy$, while at the same time the other player constantly updates $y$ to minimize $f_2(y) = -xy$.

Because $\frac{\partial f_1}{\partial x} = y$ and $\frac{\partial f_2}{\partial y} = -x$, we update $x$ with $x-\eta \cdot y$ and $y$ with $y+ \eta \cdot x$ simultaneously in one iteration, where $\eta$ is the learning rate. Once $x$ and $y$ have different signs, every following gradient update causes huge oscillation and the instability gets worse in time, as shown in Fig. 3. 

\begin{figure}[!htb]
	\centering
	\includegraphics[width=0.8\linewidth]{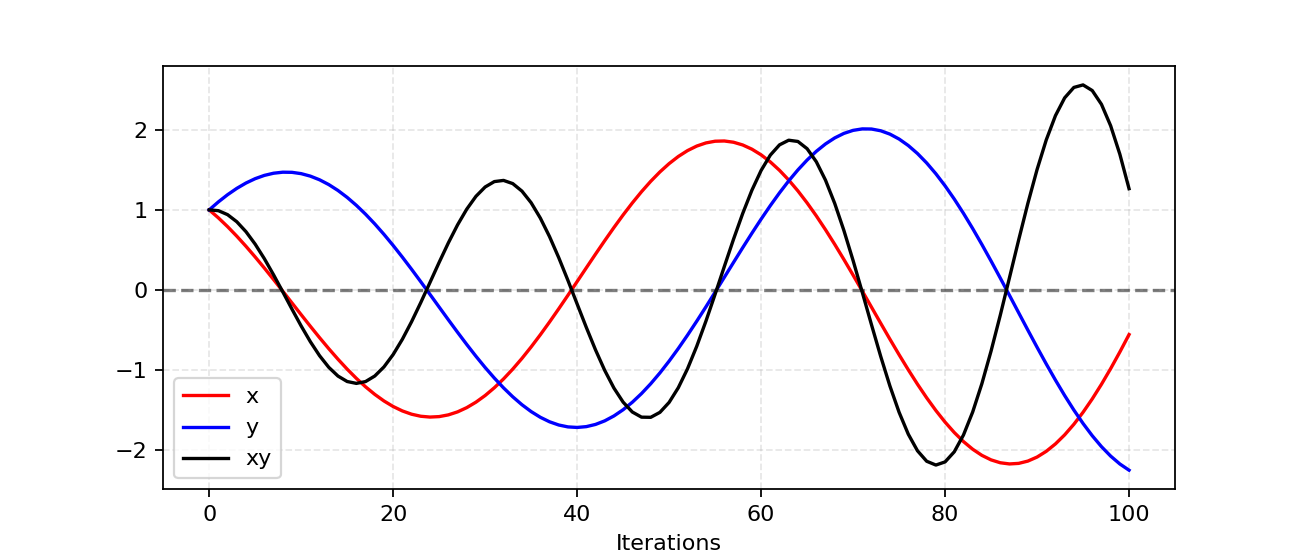}
	\caption{A simulation of our example for updating $x$ to minimize $xy$ and updating $y$ to minimize $-xy$. The learning rate $\eta = 0.1$. With more iterations, the oscillation grows more and more unstable.}
	\label{fig:fig3}
\end{figure}

\subsection{Low Dimensional Supports}
\label{sec:low_dimensional_supports}

\begin{table}[h!]
	\centering
	\begin{tabular}{c|p{10cm}} 
		\hline
		\textbf{Term} & \textbf{Explanation} \\
		\hline
		Manifold & A topological space that locally resembles Euclidean space near each point. Precisely, when this Euclidean space is of dimension $n$, the manifold is referred as $n$-manifold. \\
		Support & A real-valued function $f$ is the subset of the domain containing those elements which are not mapped to zero.\\
		\hline
	\end{tabular}
\end{table}

\cite{arjovsky2017} discussed the problem of the supports of $p_r$ and $p_g$ lying on low dimensional manifolds and how it contributes to the instability of GAN training thoroughly.

The dimensions of many real-world datasets, as represented by $p_r$, only appear to be \textit{artificially high}. They have been found to concentrate in a lower dimensional manifold. This is actually the fundamental assumption for \textit{Manifold Learning}. Thinking of the real world images, once the theme or the contained object is fixed, the images have a lot of restrictions to follow, i.e., a dog should have two ears and a tail, and a skyscraper should have a straight and tall body, etc. These restrictions keep images away from the possibility of having a high-dimensional free form.

$p_g$ lies in a low dimensional manifolds, too. Whenever the generator is asked to a much larger image like 64x64 given a small dimension, such as 100, noise variable input $z$, the distribution of colors over these 4096 pixels has been defined by the small 100-dimension random number vector and can hardly fill up the whole high dimensional space.

\begin{figure}[!htb]
	\centering
	\includegraphics[width=\linewidth]{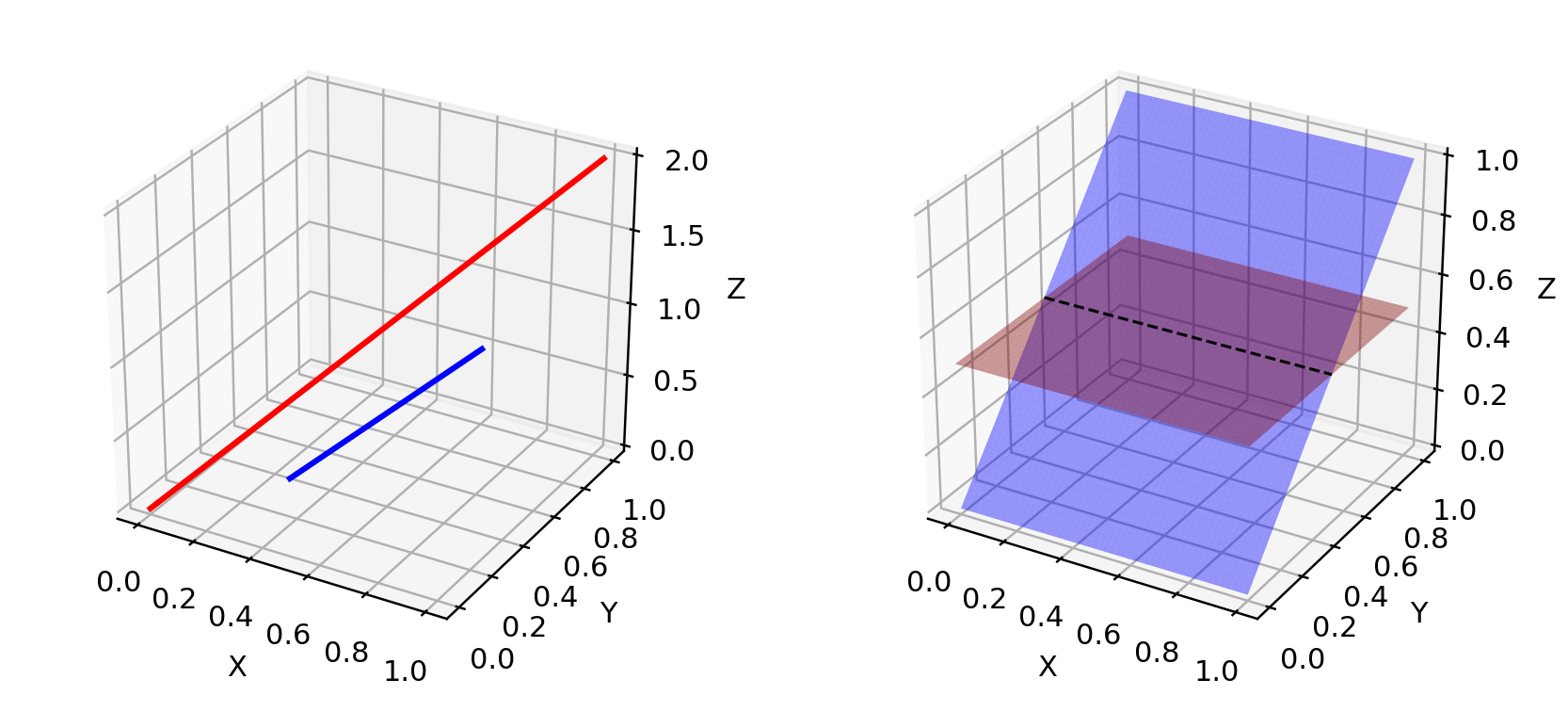}
	\caption{Low dimensional manifolds in high dimension space can hardly have overlaps. (Left) Two lines in a three-dimension space. (Right) Two surfaces in a three-dimension space.}
	\label{fig:fig4}
\end{figure}

Because both $p_g$ and $p_r$ rest in low dimensional manifolds, they are almost certainly gonna be disjoint (See Fig.~\ref{fig:fig4}). When they have disjoint supports, we are always capable of finding a perfect discriminator that separates real and fake samples 100\% correctly.~\cite{arjovsky2017}

\subsection{Vanishing Gradient}

When the discriminator is perfect, we are guaranteed with $D(x) = 1, \forall x \in p_r$ and $D(x) = 0, \forall x \in p_g$. Therefore the loss function $L$ falls to zero and we end up with no gradient to update the loss during learning iterations. Fig. 5 demonstrates an experiment when the discriminator gets better, the gradient vanishes fast.

\begin{figure}[!htb]
	\centering
	\includegraphics[width=0.6\linewidth]{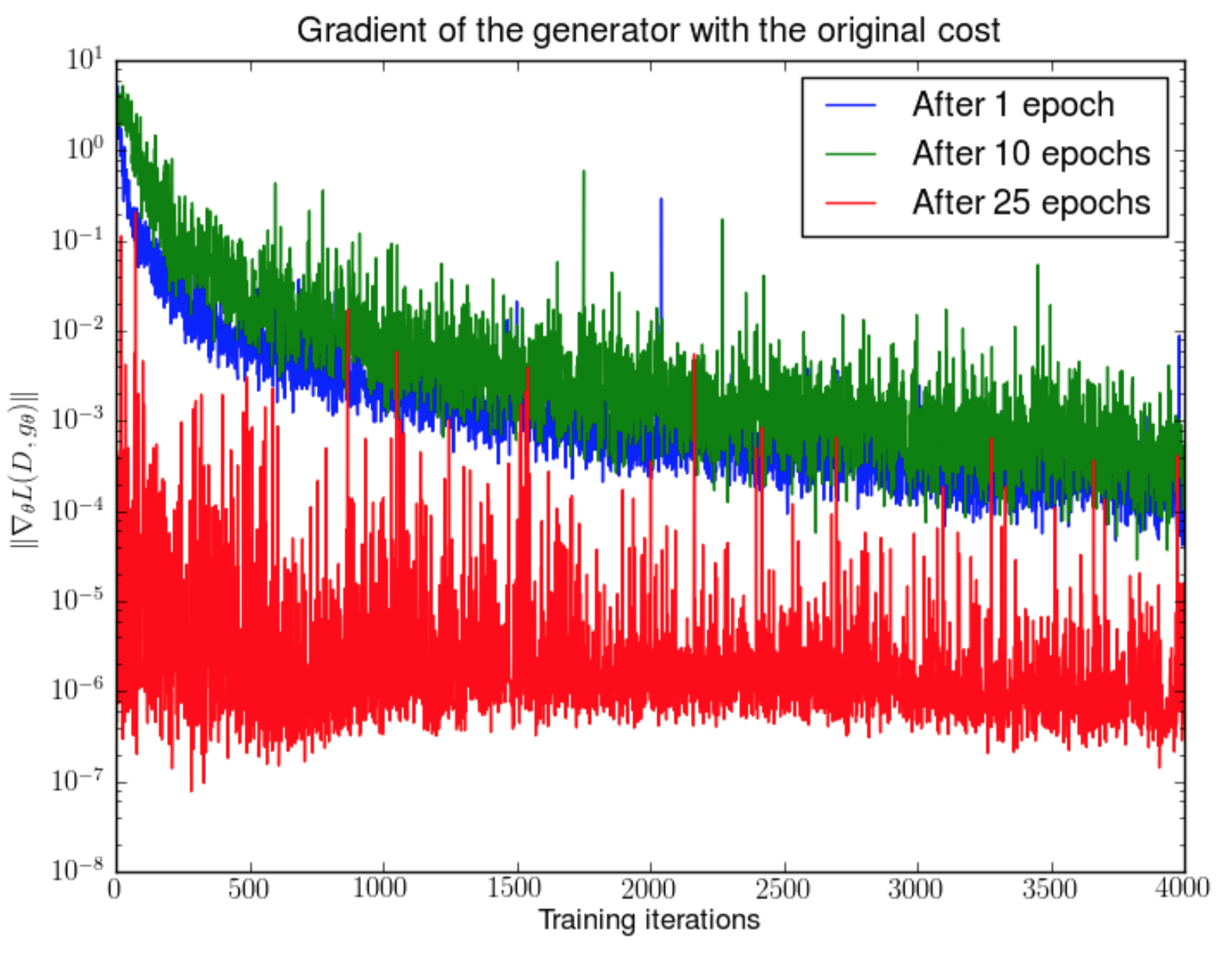}
	\caption{First, a DCGAN is trained for 1, 10 and 25 epochs. Then, with the generator \textit{fixed}, a discriminator is trained from scratch and measure the gradients with the original cost function. We see the gradient norms \textit{decay quickly} (in log scale), in the best case 5 orders of magnitude after 4000 discriminator iterations. (Image source:~\cite{arjovsky2017}).}
	\label{fig:fig5}
\end{figure}

As a result, training a GAN faces an dilemma:
\begin{itemize}
	\item If the discriminator behaves badly, the generator does not have accurate feedback and the loss function cannot represent the reality.
	\item If the discriminator does a great job, the gradient of the loss function drops down to close to zero and the learning becomes super slow or even jammed.
\end{itemize}

This dilemma clearly is capable to make the GAN training very tough.

\subsection{Mode Collapse}

During the training, the generator may collapse to a setting where it always produces same outputs. This is a common failure case for GANs, commonly referred to as \textit{Mode Collapse}. Even though the generator might be able to trick the corresponding discriminator, it fails to learn to represent the complex real-world data distribution and gets stuck in a small space with extremely low variety.

\begin{figure}[!htb]
	\centering
	\includegraphics[width=\linewidth]{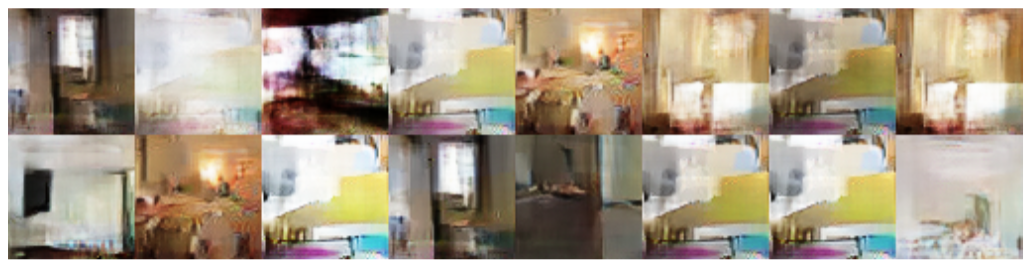}
	\caption{A DCGAN model is trained with an MLP network with 4 layers, 512 units and ReLU activation function, configured to lack a strong inductive bias for image generation. The results shows a significant degree of mode collapse. (Image source:~\cite{wgan2017}).}
	\label{fig:fig6}
\end{figure}

\subsection{Lack of a Proper Evaluation Metric}

Generative adversarial networks are not born with a good objection function that can inform us the training progress. Without a good evaluation metric, it is like working in the dark. No good sign to tell when to stop; No good indicator to compare the performance of multiple models.

\section{Improved GAN Training}

The following suggestions are proposed to help stabilize and improve the training of GANs.

First five methods are practical techniques to achieve faster convergence of GAN training~\cite{salimans2016nips}. The last two are proposed in~\cite{arjovsky2017} to solve the problem of disjoint distributions.

(1) \textbf{Feature Matching}

Feature matching suggests to optimize the discriminator to inspect whether the generator's output matches expected statistics of the real samples. In such a scenario, the new loss function is defined as $\| \mathbb{E}_{x \sim p_r} f(x) - \mathbb{E}_{z \sim p_z(z)}f(G(z)) \|_2^2 $, where $f(x)$ can be any computation of statistics of features, such as mean or median.

(2) \textbf{Minibatch Discrimination}

With minibatch discrimination, the discriminator is able to digest the relationship between training data points in one batch, instead of processing each point independently. 

In one minibatch, we approximate the closeness between every pair of samples, $c(x_i, x_j)$, and get the overall summary of one data point by summing up how close it is to other samples in the same batch, $o(x_i) = \sum_{j} c(x_i, x_j)$. Then $o(x_i)$ is explicitly added to the input of the model.

(3) \textbf{Historical Averaging}

For both models, add $ \| \Theta - \frac{1}{t} \sum_{i=1}^t \Theta_i \|^2 $ into the loss function, where $\Theta$ is the model parameter and $\Theta_i$ is how the parameter is configured at the past training time $i$. This addition piece penalizes the training speed when $\Theta$ is changing too dramatically in time.

(4) \textbf{One-sided Label Smoothing}

When feeding the discriminator, instead of providing 1 and 0 labels, use soften values such as 0.9 and 0.1. It is shown to reduce the networks' vulnerability.

(5) \textbf{Virtual Batch Normalization (VBN)}

Each data sample is normalized based on a fixed batch (\textit{"reference batch"}) of data rather than within its minibatch. The reference batch is chosen once at the beginning and stays the same through the training.

(6) \textbf{Adding Noises}

Based on the discussion in Sec.~\ref{sec:low_dimensional_supports}, we now know $p_r$ and $p_g$ are disjoint in a high dimensional space and it causes the problem of vanishing gradient. To artificially "spread out" the distribution and to create higher chances for two probability distributions to have overlaps, one solution is to add continuous noises onto the inputs of the discriminator $D$.

(7) \textbf{Use Better Metric of Distribution Similarity}

The loss function of the vanilla GAN measures the JS divergence between the distributions of $p_r$ and $p_g$. This metric fails to provide a meaningful value when two distributions are disjoint.

Wasserstein metric is proposed to replace JS divergence because it has a much smoother value space. See more in the next section.

\section{Wasserstein GAN (WGAN)}

\subsection{What is Wasserstein Distance?}

\textit{Wasserstein Distance} is a measure of the distance between two probability distributions.
It is also called \textit{Earth Mover's distance}, short for EM distance, because informally it can be interpreted as the minimum energy cost of moving and transforming a pile of dirt in the shape of one probability distribution to the shape of the other distribution. The cost is quantified by: the amount of dirt moved x the moving distance.

Let us first look at a simple case where the probability domain is discrete. For example, suppose we have two distributions $P$ and $Q$, each has four piles of dirt and both have ten shovelfuls of dirt in total. The numbers of shovelfuls in each dirt pile are assigned as follows:

\[
P_1 = 3, P_2 = 2, P_3 = 1, P_4 = 4\\
Q_1 = 1, Q_2 = 2, Q_3 = 4, Q_4 = 3
\]

\begin{figure}[!htb]
	\centering
	\includegraphics[width=\linewidth]{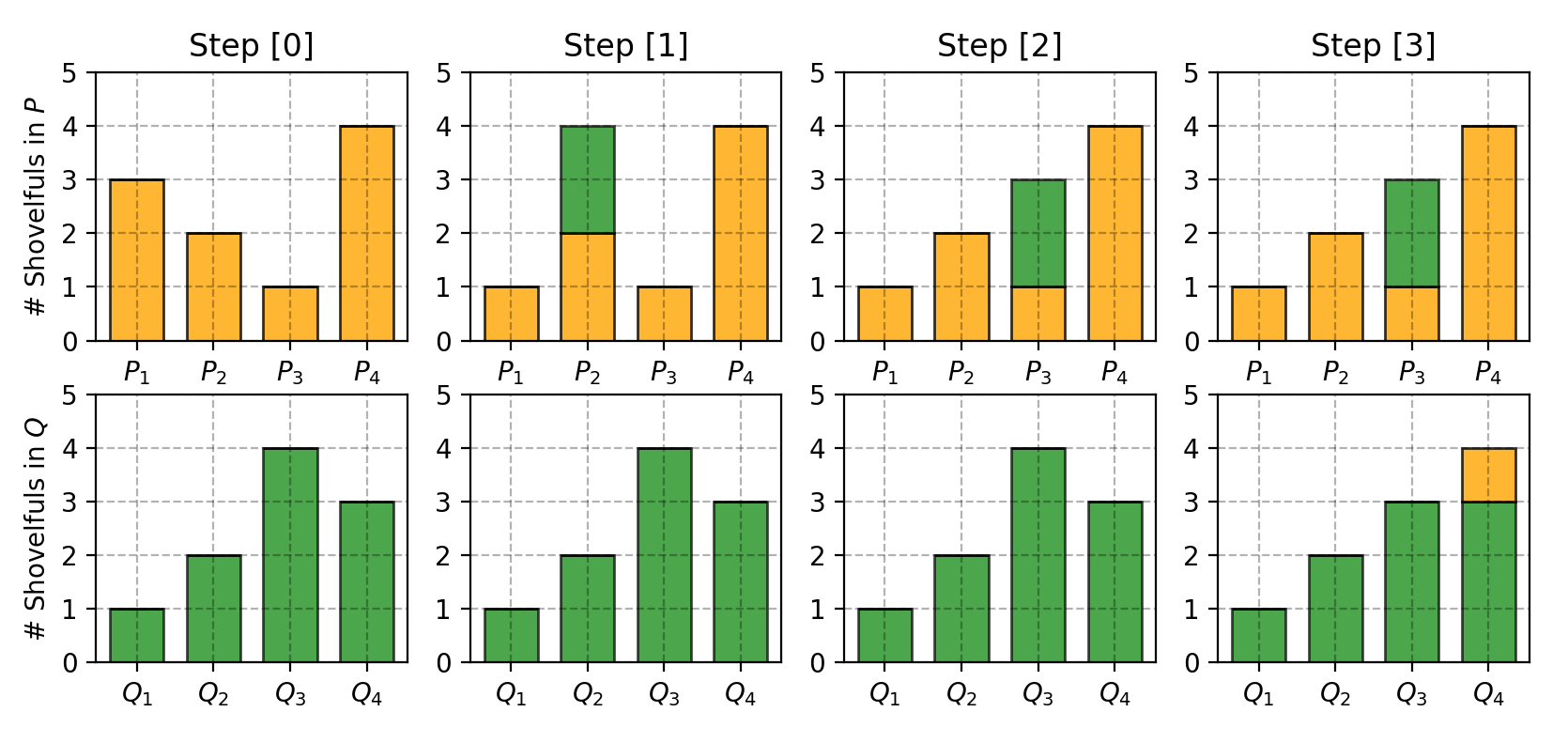}
	\caption{Step-by-step plan of moving dirt between piles in $P$ and $Q$ to make them match.}
	\label{fig:fig7}
\end{figure}

In order to change $P$ to look like $Q$, as illustrated in Fig.~\ref{fig:fig7}, we:
\begin{itemize}
	\item First move 2 shovelfuls from $P_1$ to $P_2$ => $(P_1, Q_1)$ match up.
	\item Then move 2 shovelfuls from $P_2$ to $P_3$ => $(P_2, Q_2)$ match up.
	\item Finally move 1 shovelfuls from $Q_3$ to $Q_4$ => $(P_3, Q_3)$ and $(P_4, Q_4)$ match up. 
\end{itemize}

If we label the cost to pay to make $P_i$ and $Q_i$ match as $\delta_i$, we would have $\delta_{i+1} = \delta_i + P_i - Q_i$ and in the example:

\begin{align*}
\delta_0 &= 0\\
\delta_1 &= 0 + 3 - 1 = 2\\
\delta_2 &= 2 + 2 - 2 = 2\\
\delta_3 &= 2 + 1 - 4 = -1\\
\delta_4 &= -1 + 4 - 3 = 0
\end{align*}

Finally the Earth Mover's distance is $W = \sum \vert \delta_i \vert = 5$.

When dealing with the continuous probability domain, the distance formula becomes:

\[
W(p_r, p_g) = \inf_{\gamma \sim \Pi(p_r, p_g)} \mathbb{E}_{(x, y) \sim \gamma}[\| x-y \|]
\]

In the formula above, $\Pi(p_r, p_g)$ is the set of all possible joint probability distributions between $p_r$ and $p_g$. One joint distribution $\gamma \in \Pi(p_r, p_g)$ describes one dirt transport plan, same as the discrete example above, but in the continuous probability space. Precisely $\gamma(x, y)$ states the percentage of dirt should be transported from point $x$ to $y$ so as to make $x$ follows the same probability distribution of $y$. That's why the marginal distribution over $x$ adds up to $p_g$, $\sum_{x} \gamma(x, y) = p_g(y)$ (Once we finish moving the planned amount of dirt from every possible $x$ to the target $y$, we end up with exactly what $y$ has according to $p_g$.) and vice versa $\sum_{y} \gamma(x, y) = p_r(x)$.

When treating $x$ as the starting point and $y$ as the destination, the total amount of dirt moved is $\gamma(x, y)$ and the traveling distance is $\| x-y \|$ and thus the cost is $\gamma(x, y) \cdot \| x-y \|$. The expected cost averaged across all the $(x,y)$ pairs can be easily computed as:

\[
\sum_{x, y} \gamma(x, y) \| x-y \| 
= \mathbb{E}_{x, y \sim \gamma} \| x-y \|
\]

Finally, we take the minimum one among the costs of all dirt moving solutions as the EM distance. In the definition of Wasserstein distance, the $\inf$ (infimum, also known as *greatest lower bound*) indicates that we are only interested in the smallest cost.

\subsection{Why Wasserstein is better than JS or KL Divergence?}

Even when two distributions are located in lower dimensional manifolds without overlaps, Wasserstein distance can still provide a meaningful and smooth representation of the distance in-between.

The WGAN paper exemplified the idea with a simple example.

Suppose we have two probability distributions, $P$ and $Q$:

\[
\forall (x, y) \in P, x = 0 \text{ and } y \sim U(0, 1)\\
\forall (x, y) \in Q, x = \theta, 0 \leq \theta \leq 1 \text{ and } y \sim U(0, 1)\\
\]

\begin{figure}[!htb]
	\centering
	\includegraphics[width=0.7\linewidth]{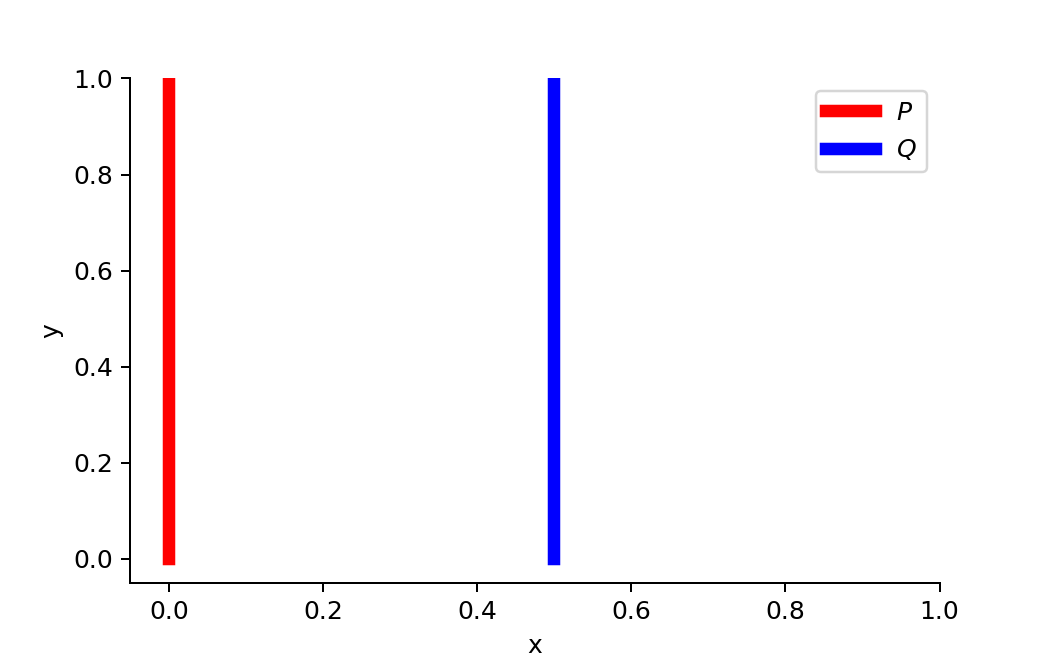}
	\caption{There is no overlap between $P$ and $Q$ when $\theta \neq 0$.}
	\label{fig:fig8}
\end{figure}

When $\theta \neq 0$:

\begin{align*}
D_{KL}(P \| Q) &= \sum_{x=0, y \sim U(0, 1)} 1 \cdot \log\frac{1}{0} = +\infty \\
D_{KL}(Q \| P) &= \sum_{x=\theta, y \sim U(0, 1)} 1 \cdot \log\frac{1}{0} = +\infty \\
D_{JS}(P, Q) &= \frac{1}{2}(\sum_{x=0, y \sim U(0, 1)} 1 \cdot \log\frac{1}{1/2} + \sum_{x=0, y \sim U(0, 1)} 1 \cdot \log\frac{1}{1/2}) = \log 2\\
W(P, Q) &= |\theta|
\end{align*}

But when $\theta = 0$, two distributions are fully overlapped:

\begin{align*}
D_{KL}(P \| Q) &= D_{KL}(Q \| P) = D_{JS}(P, Q) = 0\\
W(P, Q) &= 0 = \lvert \theta \rvert
\end{align*}

$D_{KL}$ gives us infinity when two distributions are disjoint. The value of $D_{JS}$ has sudden jump, not differentiable at $\theta = 0$. Only Wasserstein metric provides a smooth measure, which is super helpful for a stable learning process using gradient descents.

\subsection{Use Wasserstein Distance as GAN Loss Function}

It is intractable to exhaust all the possible joint distributions in $\Pi(p_r, p_g)$ to compute $\inf_{\gamma \sim \Pi(p_r, p_g)}$. Thus the authors proposed a smart transformation of the formula based on the Kantorovich-Rubinstein duality to:

\[
W(p_r, p_g) = \frac{1}{K} \sup_{\| f \|_L \leq K} \mathbb{E}_{x \sim p_r}[f(x)] - \mathbb{E}_{x \sim p_g}[f(x)]
\]

where $\sup$ (supremum) is the opposite of $inf$ (infimum); we want to measure the least upper bound or, in even simpler words, the maximum value.

\subsubsection{Lipschitz Continuity}

The function $f$ in the new form of Wasserstein metric is demanded to satisfy $\| f \|_L \leq K$, meaning it should be \textit{K-Lipschitz continuous}.

A real-valued function $f: \mathbb{R} \rightarrow \mathbb{R}$ is called $K$-Lipschitz continuous if there exists a real constant $K \geq 0$ such that, for all $x_1, x_2 \in \mathbb{R}$,

\[
\lvert f(x_1) - f(x_2) \rvert \leq K \lvert x_1 - x_2 \rvert
\]

Here $K$ is known as a Lipschitz constant for function $f(.)$. Functions that are everywhere continuously differentiable is Lipschitz continuous, because the derivative, estimated as $\frac{\lvert f(x_1) - f(x_2) \rvert}{\lvert x_1 - x_2 \rvert}$, has bounds. However, a Lipschitz continuous function may not be everywhere differentiable, such as $f(x) = \lvert x \rvert$.

Explaining how the transformation happens on the Wasserstein distance formula is worthy of a long post by itself, so I skip the details here. If you are interested in how to compute Wasserstein metric using linear programming, or how to transfer Wasserstein metric into its dual form according to the Kantorovich-Rubinstein Duality, read this awesome \href{https://vincentherrmann.github.io/blog/wasserstein/}{post}.

\subsubsection{Wasserstein Loss Function}

Suppose this function $f$ comes from a family of K-Lipschitz continuous functions, $\{ f_w \}_{w \in W}$, parameterized by $w$. In the modified Wasserstein-GAN, the "discriminator" model is used to learn $w$ to find a good $f_w$ and the loss function is configured as measuring the Wasserstein distance between $p_r$ and $p_g$.

\[
L(p_r, p_g) = W(p_r, p_g) = \max_{w \in W} \mathbb{E}_{x \sim p_r}[f_w(x)] - \mathbb{E}_{z \sim p_r(z)}[f_w(g_\theta(z))]
\]

Thus the "discriminator" is not a direct critic of telling the fake samples apart from the real ones anymore. Instead, it is trained to learn a $K$-Lipschitz continuous function to help compute Wasserstein distance. As the loss function decreases in the training, the Wasserstein distance gets smaller and the generator model's output grows closer to the real data distribution.

One big problem is to maintain the $K$-Lipschitz continuity of $f_w$ during the training in order to make everything work out. The paper presents a simple but very practical trick: After every gradient update, clamp the weights $w$ to a small window, such as $[-0.01, 0.01]$, resulting in a compact parameter space $W$ and thus $f_w$ obtains its lower and upper bounds to preserve the Lipschitz continuity.

\begin{figure}[!htb]
	\centering
	\includegraphics[width=0.85\linewidth]{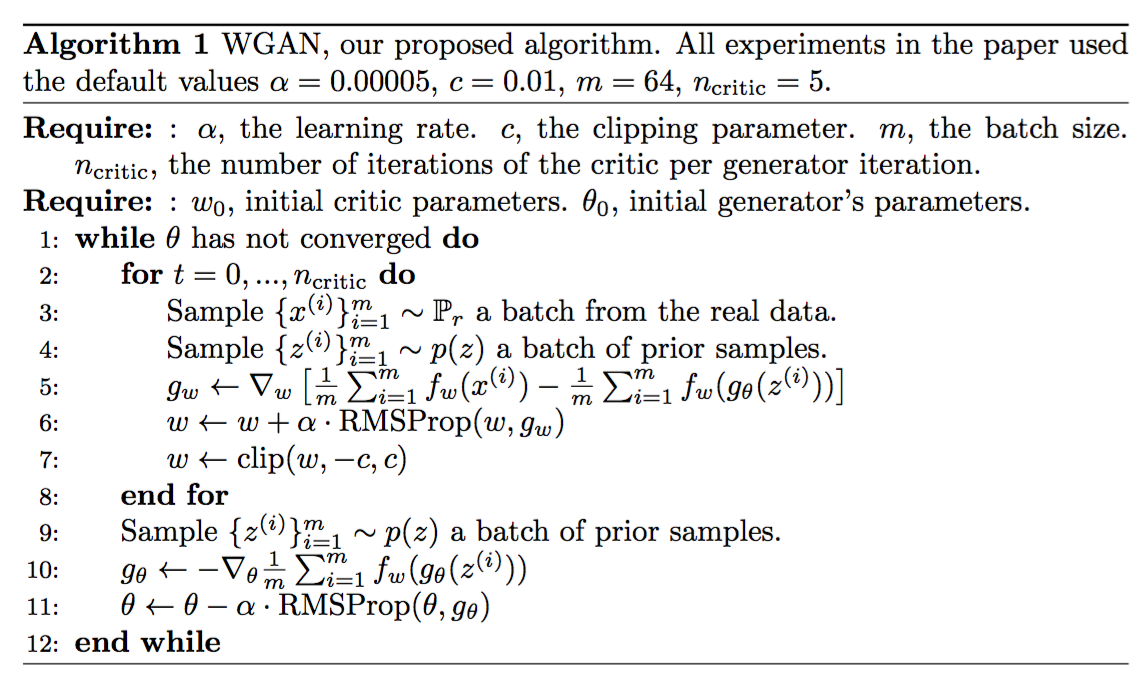}
	\caption{Algorithm of Wasserstein generative adversarial network. (Image source:~\cite{wgan2017})}
	\label{fig:fig9}
\end{figure}

Compared to the original GAN algorithm, the WGAN undertakes the following changes:
\begin{itemize}
	\item After every gradient update on the critic function, clamp the weights to a small fixed range, $[-c, c]$.
	\item Use a new loss function derived from the Wasserstein distance, no logarithm anymore. The "discriminator" model does not play as a direct critic but a helper for estimating the Wasserstein metric between real and generated data distribution.
	\item Empirically the authors recommended RMSProp optimizer on the critic, rather than a momentum based optimizer such as Adam which could cause instability in the model training. I haven't seen clear theoretical explanation on this point through.
\end{itemize}

Sadly, Wasserstein GAN is not perfect. Even the authors of the original WGAN paper mentioned that \textit{"Weight clipping is a clearly terrible way to enforce a Lipschitz constraint"}. WGAN still suffers from unstable training, slow convergence after weight clipping (when clipping window is too large), and vanishing gradients (when clipping window is too small).

Some improvement, precisely replacing weight clipping with \textit{gradient penalty}, has been discussed in~\cite{wgan2017improve}.

\bibliographystyle{plain}
%\bibliography{../references}

\begin{thebibliography}{1}
	
	\bibitem{gan2014}
	Ian Goodfellow, Jean Pouget-Abadie, Mehdi Mirza, Bing Xu, David Warde-Farley,
	Sherjil Ozair, Aaron Courville, and Yoshua Bengio.
	\newblock Generative adversarial nets.
	\newblock In {\em NIPS}, pages 2672--2680. 2014.
	
	\bibitem{gan2015train}
	Ferenc Huszár.
	\newblock How (not) to train your generative model: Scheduled sampling,
	likelihood, adversary?
	\newblock {\em arXiv:1511.05101}, 2015.
	
	\bibitem{salimans2016nips}
	Tim Salimans, Ian Goodfellow, Wojciech Zaremba, Vicki Cheung, Alec Radford, and
	Xi~Chen.
	\newblock Improved techniques for training gans.
	\newblock {\em NIPS}, 2016.
	
	\bibitem{arjovsky2017}
	Martin Arjovsky and Léon Bottou.
	\newblock Towards principled methods for training generative adversarial
	networks.
	\newblock {\em ICML}, 2017.
	
	\bibitem{wgan2017}
	Martin Arjovsky, Soumith Chintala, and Léon Bottou.
	\newblock Wasserstein gan.
	\newblock {\em arXiv:1701.07875}, 2017.
	
	\bibitem{wgan2017improve}
	Ishaan Gulrajani, Faruk Ahmed, Martin Arjovsky, Vincent Dumoulin, and Aaron
	Courville.
	\newblock Improved training of wasserstein gans.
	\newblock {\em arXiv:1704.00028}, 2017.
	
\end{thebibliography}

\end{document}